\documentclass[sigconf]{acmart}
\AtBeginDocument{%
  }
\renewcommand\footnotetextcopyrightpermission[1]{}
\setcopyright{acmlicensed}
\copyrightyear{2025}
\acmYear{2025}
\acmDOI{XXXXXXX.XXXXXXX}
\acmConference[Conference ACM MM '2025]{Make sure to enter the correct
  conference title from your rights confirmation email}{October 27--31,
  2025}{Dublin, Ireland}

\usepackage{pifont}
\usepackage{algorithm}
\usepackage{algpseudocode}
\usepackage{multicol, multirow}
\usepackage{makecell} 
\usepackage{graphicx} 
\usepackage{array}
\usepackage{pifont}
\usepackage{xcolor}

\begin{document}

\title{Try Harder: Hard Sample Generation and Learning for Clothes-Changing Person Re-ID}


\author{Hankun Liu}
\authornote{These two authors contributed equally to this work.}  
\affiliation{%
  \institution{School of Computer Science and Engineering, Beihang University}
  \country{China}
}
\email{lhk2783497478@buaa.edu.cn}

\author{Yujian Zhao}
\authornotemark[1]  
\affiliation{%
   \institution{School of Artificial Intelligence, Beihang University} 
   \institution{Zhongguancun Academy}
   \country{China}
}
\email{yjzhao1019@buaa.edu.cn}

\author{Guanglin Niu}
\authornote{Corresponding author.  This work was supported by the National Natural Science Foundation of China (No. 62376016).}

\affiliation{%
  \institution{School of Artificial Intelligence, Beihang University}
  \country{China}
}
\email{beihangngl@buaa.edu.cn}  








\renewcommand{\shortauthors}{Hankun Liu et al.}

\begin{abstract}

  Hard samples pose a significant challenge in person re-identification (ReID) tasks, particularly in clothing-changing person Re-ID (CC-ReID). Their inherent ambiguity or similarity, coupled with the lack of explicit definitions, makes them a fundamental bottleneck. These issues not only limit the design of targeted learning strategies but also diminish the model's robustness under clothing or viewpoint changes.
  In this paper, we propose a novel multimodal-guided Hard Sample Generation and Learning (HSGL) framework, which is the first effort to unify textual and visual modalities to explicitly define, generate, and optimize hard samples within a unified paradigm. HSGL comprises two core components: (1) Dual-Granularity Hard Sample Generation (DGHSG), which leverages multimodal cues to synthesize semantically consistent samples, including both coarse- and fine-grained hard positives and negatives for effectively increasing the hardness and diversity of the training data. (2) Hard Sample Adaptive Learning (HSAL), which introduces a hardness-aware optimization strategy that adjusts feature distances based on textual semantic labels, encouraging the separation of hard positives and drawing hard negatives closer in the embedding space to enhance the model’s discriminative capability and robustness to hard samples.
  Extensive experiments on multiple CC-ReID benchmarks demonstrate the effectiveness of our approach and highlight the potential of multimodal-guided hard sample generation and learning for robust CC-ReID. Notably, HSAL significantly accelerates the convergence of the targeted learning procedure and achieves state-of-the-art performance on both PRCC and LTCC datasets. The code is available at \url{https://github.com/undooo/TryHarder-ACMMM25}.
    
\end{abstract}




\maketitle

\begin{figure}[h]
  \centering
  \includegraphics[width=\linewidth]{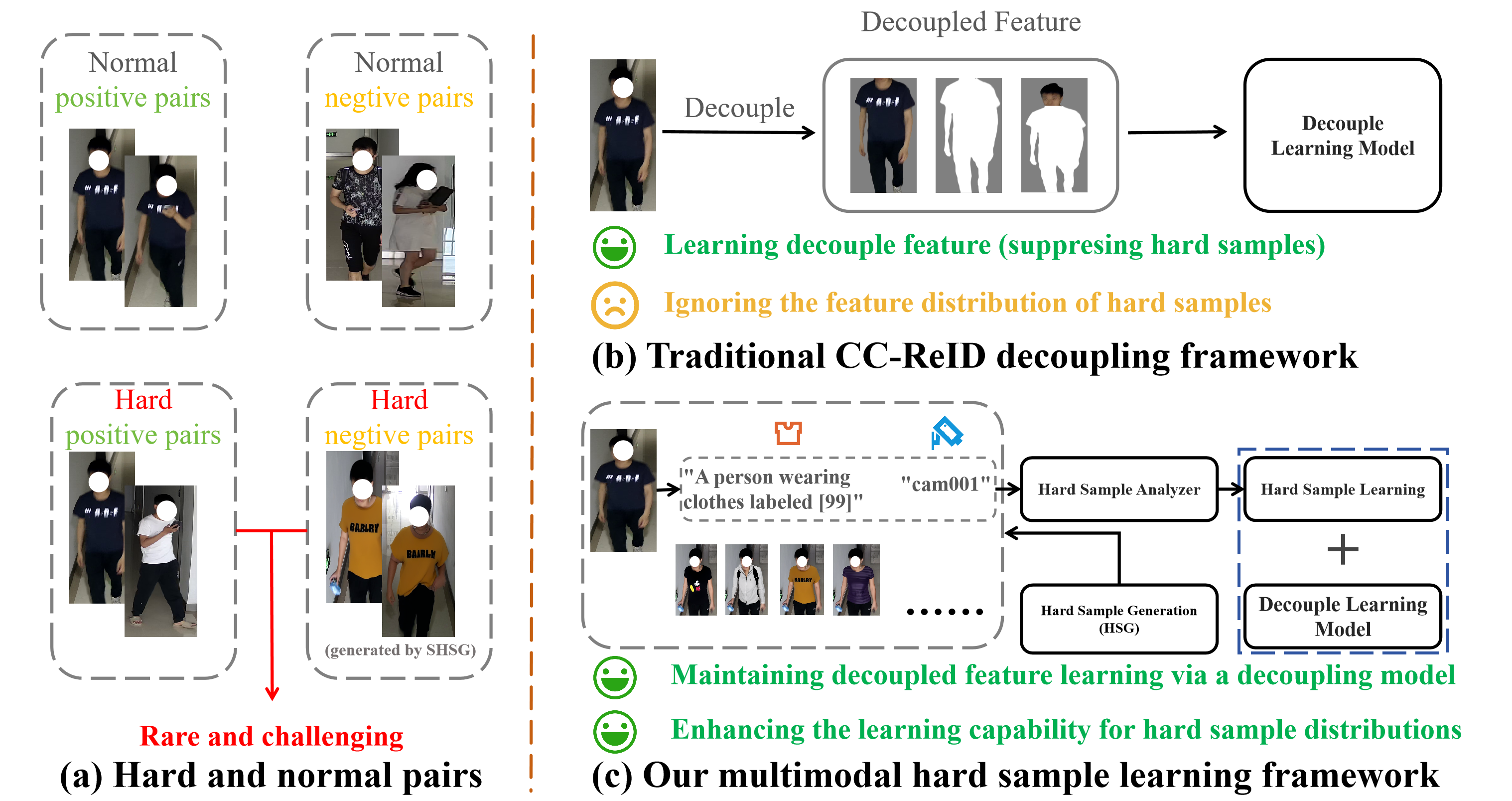}
  \caption{Conceptual comparison of hard samples and solutions in CC-ReID.
        (a) Hard samples definition.
        (b) Conventional approaches focus on feature disentanglement or static metric learning. 
        (c) Our multimodal hard sample learning framework.
  }
  \label{intro}
\end{figure}

\section{Introduction}

Cloth-changing person re-identification (CC-ReID) aims to match pedestrian identities across clothing variations, playing a pivotal role in intelligent surveillance and cross-camera tracking. Particularly, even many well-performing traditional person re-identification (ReID) models such as~\cite{TransReID, ye2021deep, leng2019survey} struggle with cloth-changing scenarios. Two key challenges in CC-ReID are the lack of effective supervision strategies specific to hard samples together with the scarcity of high-quality hard samples, both of which limit the model's capability of learning robust identity representations amidst substantial appearance changes.


\textbf{Specifically, existing studies lack precise definitions of hard samples and corresponding learning strategies.} Due to the absence of an explicit a priori definition, previous methods typically rely on feature space heuristics to identify hard samples \cite{yu2019unsupervised, CAL, AIM}. However, such representations are often unreliable in the early stages of training, leading to inaccurate identification and suboptimal gradient signals. To mitigate this, many approaches resort to feature decoupling techniques in an attempt to alleviate the ambiguity introduced by entangled appearance and identity cues \cite{IFD, FIRe, CSSC}. To address these challenges, we first establish explicit definitions of hard samples by leveraging textual clothing labels and camera tags in mainstream CC-ReID datasets: cross-clothing/cross-viewpoint samples of the same individual are defined as hard positive samples, while same-clothing samples of different identities become hard negative samples, as shown in Fig.~\ref{intro}(a). Based on this foundation, we propose Hard Sample Adaptive Learning (HSAL), a plug-and-play supervised learning framework (Fig.~\ref{intro}(c)) that dynamically adjusts feature space distance distributions. This mechanism enhances metric learning's sensitivity and intensity toward hard samples, thereby optimizing the learned distribution in feature space through the targeted learning.

\textbf{Nevertheless, the scarcity of hard samples also persists as a significant challenge}. Most of CC-ReID datasets contain merely limited clothing sets per ID (2--5 outfits)~\cite{PRCC, LTCC}, causing insufficient hard positive samples for CC-ReID tasks. A straightforward solution in our paper is developing a Coarse-grained Hard Positive Sample Generation (CHPSG) using diffusion models. However, the greater challenge of generating hard negative samples is that all the current real-world datasets cannot provide shared clothing labeling across IDs and therefore lack hard negative samples that can be leveraged for model training. To handle this issue, we design a Semantic-preserving Hard Sample Generation (SHSG) framework that synthesizes high-quality hard samples through curated ID images and clothing libraries, leveraging multimodal cloth-changing models to generate hard samples with unified cloth labels, effectively expanding the scarce sample space.

In summary, this research constructs a comprehensive framework for hard sample perception, generation, and learning in CC-ReID via multimodal generative models and information. We design: 1) explicit hard sample definitions using textual labels; 2) a complete hard sample generation pipeline enabling both coarse-grained (CHPSG) and fine-grained (SHSG) synthesis;  and 3) Hard Sample Adaptive Learning (HSAL) for effective sample utilization.

Our principal contributions are as follows:
\begin{itemize}
    \item To the best of our concern, this is the first effort to propose a novel multimodal-guided framework for explicit definition, generation, and learning of hard samples for CC-ReID.
    \item A dual-stage hard sample generation framework combining coarse-grained/large-scale hard positive sample generation (CHPSG) with fine-grained/semantically-rich hard sample generation (SHSG), solving dataset scarcity issues and facilitating targeted learning.
    \item The HSAL framework implements plug-and-play hard sample learning to enhance hard sample discrimination while preserving the original representation capacity.
    
    \item Extensive experiments show that incorporating only 18–20\% generated hard samples into training, in conjunction with HSAL, yields 3.5–4.9\% Rank-1 improvements even with just 1/10 of the training iterations. Combined with coarse-to-fine learning, our model achieves SOTA performance on PRCC and LTCC, with performance gains of +8.9\% Rank-1 and +11.5\% mAP on PRCC.
\end{itemize}

\section{Related Work}

\subsection{Cloth-changing Person Re-identification}

Cloth-changing Person Re-identification (CC-ReID) is a challenging variant of the person re-identification (ReID) task, where the clothing appearance of individuals can change across different camera views. Unlike traditional ReID methods that primarily rely on clothing-based visual cues \cite{Pose2ID, TransReID, ye2021deep, leng2019survey}, CC-ReID focuses on learning identity-discriminative representations that are invariant to clothing variations. There are three mainstream approaches to address this challenge: hard sample mining and learning\cite{yu2019unsupervised, CHEN2020259, 9035458, zhao2024clip}, clothing-invariant feature extraction\cite{cui2023dcr, liu2023dual, Cloth-generalized, peng2024masked, CCFA, FSAM, yang2023win, guo2023semantic, TransReID, jia2022complementary}, and multimodal auxiliary feature integration\cite{GI-ReID, bansal2022cloth, li2023depth, tu2024clothing, nguyen2025ag, 3DInvar, 3DSL, nguyen2024temporal, wang2023exploring, han20223d, liu2024distilling}. 

\textbf{Hard sample mining and learning.} Hard samples play a crucial role in CC-ReID, and their presence often degrades the retrieval performance. Thus mining and learning hard samples becomes a valuable research direction in CC-ReID. However, there has been limited work related to it in recent years. In \cite{yu2019unsupervised}, the soft multilabel-guided hard negative mining method is proposed to learn a discriminative embedding across the different camera views. AHSM \cite{CHEN2020259} proposes an adaptive hard sample mining algorithm for training a robust ReID model by calculating the hard level. LoopNet \cite{9035458} proposes a listwise ranking network that can be used to choose the hard positive and negative samples globally and introduces a multiplet loss to utilize hard samples better. CLIP-DFGS \cite{zhao2024clip} proposes a hard sample mining method to offer sufficiently challenging samples to enhance CLIP’s ability to extract fine-grained features and improve the model’s ability to differentiate between individuals. Despite some previous work, the mining of hard samples relies on feature representation and hence uncertainty, which makes it difficult to learn effectively. In this paper, we propose the definition of the hard examples and introduce a learning method called HSAL that can provide effective supervised signals for model learning.

\textbf{Cloth-irrelevant feature extraction.} Aiming to build robust identity representations invariant to clothing changes, researchers propose many innovative approaches to extract cloth-irrelevant features\cite{cui2023dcr, liu2023dual, Cloth-generalized, peng2024masked, CCFA, FSAM, yang2023win, guo2023semantic, TransReID, jia2022complementary}. IFD \cite{IFD} introduces an identity-aware feature decoupling learning framework that consists of a dual-stream identity-attention model to focus comprehensively on the regions containing distinctive identity information. CAL \cite{CAL} proposes a clothes-based adversarial loss to mine clothes-irrelevant features. AIM \cite{AIM} introduces a causality-based auto-intervention model to capture clothing bias and ID clues separately and strip clothing inference from ID representation learning to simulate the entire intervention process. FIRe$^2$ \cite{FIRe} proposes a fine-grained representation and recomposition framework to explore the fine-grained information of each person through clustering, and then facilitate the model learning with fine-grained representation learning. In this paper, we propose a Hard Sample Adaptive Learning (HSAL) that can dynamically adjust feature distance to extract discriminative cloth-irrelevant features.

\textbf{Multimodal auxiliary features.} Given the limitations of purely RGB-based models, researchers have incorporated additional modalities information, such as gait \cite{GI-ReID, bansal2022cloth, li2023depth, tu2024clothing}, 3D shape \cite{nguyen2025ag, 3DInvar, 3DSL, nguyen2024temporal, wang2023exploring, han20223d, liu2024distilling}, contour sketches \cite{yang2019person, chen2021deep}, skeleton \cite{PSCR, joseph2025clothes} and text \cite{differ, IRM} to enhance identity recognition. The fusion of multimodal information can lead to more accurate feature representation for models. GI-ReID \cite{GI-ReID} introduces gait recognition as an auxiliary modal to drive the Image ReID model to learn cloth-agnostic representations by leveraging personal unique and cloth-independent gait information. 3DInvarReID \cite{3DInvar} reconstructs accurate 3D clothed body shapes and uses them as additional modal information to assist the model in learning accurate feature representation. DIFFER \cite{differ} proposes a novel adversarial method called disentangle identity features from entangled representations that leverage textual descriptions to disentangle identity features. In our proposed HSAL, we accurately classify hard samples and force the model to learn robust features by utilizing visual and textual information. 
\subsection{Generation Model for ReID}
As the field of artificial intelligence continues to grow, valuable data becomes one of the most important parts of the deep learning task. For the CC-ReID domain, the challenge of limited high-quality data across clothing variations has prompted the exploration of synthetic data generation techniques to enhance model generalization \cite{ pose_generation, liu2024cloth, zhang2024open}. Many reasearchers utilize game engine \cite{zhang2021unrealperson, CCUP}, GAN \cite{goodfellow2014generative, deng2018image, deng2018similarity, zhong2018camera, zheng2019joint} and Diffusion model \cite{Pose2ID,zhang2024infiniteperson} to generate more diverse data. In this work, based on the diffusion and vision-language model, we introduce a multimodal dual-granularity hard sample generation method called DGHSG to drive model learning more accurate features. 

\textbf{Diffusion and  Try-on/off.} Diffusion models \cite{ho2020denoising, song2019generative, croitoru2023diffusion, rombach2022high, sohl2015deep} have recently emerged as the most powerful family of generative models \cite{zhu2023tryondiffusion}. The diffusion model is a deep generative model based on two stages, a forward diffusion stage and a reverse diffusion stage. In order to synthesize desired data, the conditional diffusion model can receive multimodal information, such as images \cite{saharia2022palette}, texts \cite{saharia2022photorealistic} and segmentation masks \cite{wu2023diffumask}, and incorporate it into the model through the attention mechanism. Try-on aims to synthesize realistic images of a person wearing a target garment, which has attracted increasing attention in computer vision due to its wide applications in e-commerce, fashion, and human modeling \cite{choi2021viton, han2018viton}. And try-off focuses on generating high-fidelity tiled garment images from human-worn garment inputs \cite{xarchakos2024tryoffanyone}. Recently, diffusion models have shown great potential in try-on and try-off by generating more photorealistic and identity-preserving results \cite{gou2023taming, yang2024texture, kim2024stableviton, zeng2024cat}. Due to their inherent capabilities to model both clothing appearance and human body structure, we introduce them to the CC-ReID task. To the best of our knowledge, this is the first attempt to do so. In our work, we import the try-on \cite{choi2024improving} and try-off \cite{xarchakos2024tryoffanyone} to generate diverse and numerous both coarse and fine-grained hard positive/negative samples, providing sufficient samples for the model.

\textbf{Multimodal data generation.} Multimodal data generation seeks to synthesize data across different modalities, such as image \cite{yang2023paint}, text \cite{ramesh2022hierarchical}, audio \cite{borsos2023audiolm}, video \cite{sun2019videobert} and pose \cite{pose_generation}, enabling richer representations and better cross-domain understanding. For the ReID domain, similar work is emerging. Pose2ID \cite{Pose2ID} proposes an identity-guided pedestrian generation paradigm that leverages different pose information to synthesize more data. \cite{zhang2025cloth} employs a cloth-changing scheme based on the stable diffusion model to generate consistent clothing for pedestrians. DLCR \cite{siddiqui2024dlcr} implements a text-guided diffusion approach to synthesize multiple images of a person with different clothes. In contrast, we fuse information from both text and image modalities and augment the original trainset with coarse-grained and fine-grained information, allowing the model to learn more accurate generalizable distributions.


\section{Methodology}
\label{sec:method}
\begin{figure*}[t]
  \centering
  \includegraphics[width=\textwidth]{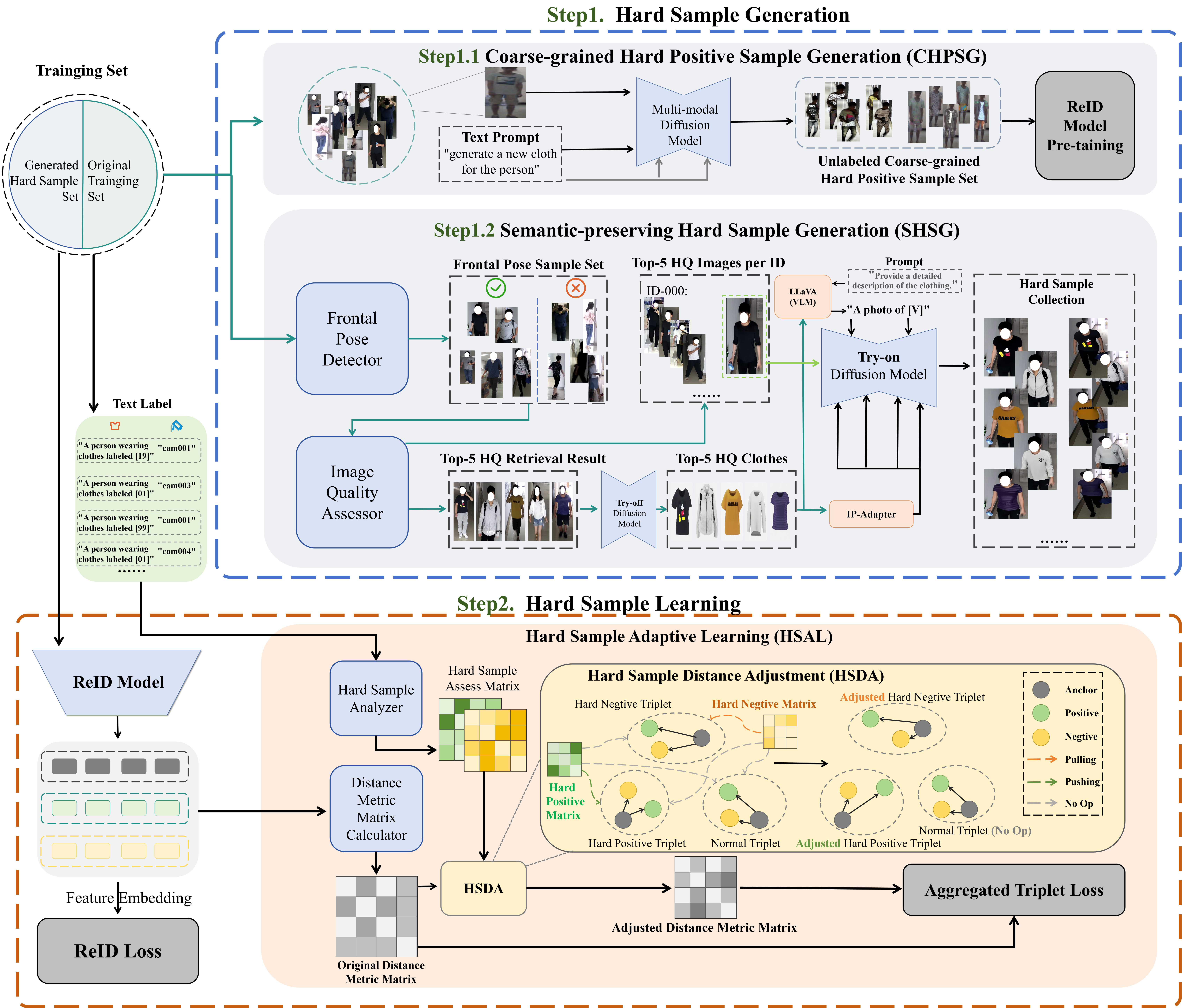}
  \caption{Overview of the proposed Hard Sample Generation And Learning(HSGL) framework. The architecture comprises three key components: 
    (1) Coarse-grained Hard Positive Sample Generation (CHPSG): A diffusion-based pipeline for large-scale generation of hard positive samples via prompt-guided clothing replacement. 
    (2) Semantic-preserving Hard Sample Generation (SHSG): A multimodal generation framework that synthesizes high-fidelity hard samples while preserving identity semantics through garment-text alignment. 
    (3)Hard Sample Adaptive Learning (HSAL): An adaptive metric learning scheme that dynamically adjusts feature space topology based on textual clothing/viewpoint annotations and visual semantics. 
  }
  \label{frame}
\end{figure*}
To the best of our knowledge, existing CC-ReID literature lacks dedicated supervised learning schemes for hard samples. To address this gap, we propose three key innovations: 1) text-guided hard sample definition, 2) dual-granularity hard sample generation, and 3) hard sample adaptive learning. Figure~\ref{frame} illustrates the overall architecture of our framework.

\subsection{Hard Sample Definition}
\label{sec:hard_def}

The absence of standardized definitions for hard samples has hindered targeted supervision algorithms. We pioneer explicit hard sample definitions using textual annotations, with core criteria focusing on clothing and viewpoint variations. Formally, given a pedestrian dataset $\mathcal{D} = \{(\mathbf{I}_i, y_i, c_i, v_i)\}_{i=1}^N$ where $y_i \in \{1,...,C\}$ denotes identity, $c_i$ clothing label, and $v_i$ viewpoint label, we define:
\begin{equation}
\begin{aligned}
    & \text{Hard Positive Pair: } &(y_i = y_j) &\land [(c_i \neq c_j) \lor (v_i \neq v_j)] \\
    & \text{Hard Negative Pair: } &(y_i \neq y_j) &\land (c_i = c_j)
\end{aligned}
\end{equation}

Notably, existing CC-ReID datasets contain limited hard negatives due to unique clothing per identity, and sparse hard positives (typically 2-5 outfits per identity)\cite{PRCC, LTCC}. This motivates our hard sample generation strategy.

\subsection{Dual-Granularity Hard Sample Generation}
\label{sec:gen}

In the era of deep learning, both the quantity and quality of data play a crucial role in determining model performance\cite{krizhevsky2012imagenet}. To address this, we propose a dual-granularity hard sample generation framework, comprising coarse-grained, large-scale hard positive sample generation and fine-grained, semantically rich hard positive and negative sample synthesis.  

As illustrated in Fig.~\ref{frame}, we utilize a multimodal virtual try-on model to generate high-quality hard samples. However, its low efficiency and strict input quality requirements make it unsuitable for large-scale generation of high-quality hard positive and negative samples. To overcome this limitation, we employ a text-guided diffusion model for large-scale, coarse-grained hard positive sample generation. By integrating these two approaches, we ensure a sufficient supply of hard samples to enhance model training.

\subsubsection{\textbf{Semantic-Preserving Hard Sample Generation (SHSG)}}  

Before performing clothing extraction and virtual try-on generation, it is essential to select pedestrian images with relatively frontal poses and high image quality\cite{gou2023taming, yang2024texture, kim2024stableviton, zeng2024cat}. This requirement arises because virtual try-on models are typically trained on high-resolution images of forward-facing models, and their performance degrades significantly when applied to low-quality pedestrian images. To address this issue, we design a frontal pose detector and an image quality evaluator to filter the samples.

\textbf{Frontal Pose Detector.} The frontal pose detector leverages MediaPipe\cite{lugaresi2019mediapipe} to detect human key points and determine whether the face is oriented forward using a comprehensive evaluation strategy. First, we ensure the reliability of fundamental facial features by requiring the visibility of the nose and eye key points to exceed 0.7. Next, geometric constraints are applied by calculating the height difference between the left and right eyes to assess horizontal alignment, while also ensuring that the interocular distance is not excessively small to avoid side-profile images. Additionally, symmetry is analyzed, as a truly frontal-facing pose should exhibit similar visibility for both ears. If one ear is significantly more visible than the other, the image is classified as a side-profile pose. By integrating these criteria, we compute a comprehensive score to determine the final classification. The relevant constraints are formulated as follows:
\begin{equation}
\small
\begin{cases} 
 \text{Visibility: } \max(k_{\text{nose}}, k_{\text{eyes}}) > 0.7 \\
 \text{Geometric: } |y_{\text{left eye}} - y_{\text{right eye}}| < \epsilon_y \\
 \text{Symmetry: } |v_{\text{left ear}} - v_{\text{right ear}}| < \epsilon_v
 \end{cases}
\end{equation}

\textbf{Image Quality Assessor.} The image quality assessor evaluates images based on resolution and sharpness. Resolution is measured by calculating the total number of pixels ($R = w \times h$), while sharpness is assessed using the variance of the Laplacian ($S = \text{Var}(\nabla^2 \mathbf{I})$)\cite{belkin2003laplacian}, enabling effective identification of blurred and clear images. To ensure high-quality inputs, the selection process prioritizes images with higher resolution and minimal blurriness.

\textbf{Try-off model.} Differences in clothing styles across datasets can affect the consistency of image appearance. To ensure a uniform visual style, we employ a high-quality sample selector to identify the top five highest-quality images from the entire dataset. These selected images are then processed using the Try-Off model to extract 2D clothing representations, which are stored in a clothing database. Specifically, we utilize TryOffAnyone\cite{xarchakos2024tryoffanyone}, a model that leverages DiT and U-Net to learn the relationship between human figures and clothing images, enabling accurate clothing extraction from individuals.

\textbf{Multimodal Try-on model.} Generating uniformly dressed pedestrian images allows for simultaneous expansion of hard positive and negative samples. Based on a high-quality sample selection filter, we select $n$ photographs with optimal evaluation results for each identity, combined with $m$ garments from our clothing repository as input, generating $m \times n$ virtual try-on images by IDM-VTON\cite{choi2024improving}, which has shown to be one of the most satisfactory open-source virtual try-on models currently available. 

Through our generation pipeline, let $C = |\mathcal{C}|$ denote the total number of distinct identities, $m$ the size of the clothing repository, and $n$ the number of selected images per identity. For each identity $i \in \{1,...,C\}$, let $K_i$ represent its original image count. The framework produces:
\begin{equation}
\begin{aligned}
    N_{\text{HP}} &= \sum_{i=1}^{C} mn(K_i + mn - 1) \\
    N_{\text{HN}} &= mnC \cdot n(C - 1)
\end{aligned}
\end{equation}
where $N_{\text{HP}}$ denotes hard positive and $N_{\text{HN}}$ indicates hard negative sample pairs with the following characteristics:
\begin{itemize}
    \item $N_{\text{HP}}$ counts cross-garment/cross-view pairs within identities
    \item $N_{\text{HN}}$ counts same-garment pairs across different identities
\end{itemize}

\subsubsection{\textbf{Coarse-Grained Hard Positive Sample Generation (CHPSG)}}

Due to the reliance of the virtual try-on model on high-quality pedestrian images, its performance decreases significantly when handling low-quality data or unfavorable viewing angles. In addition, because the model requires about 15 seconds per image even when using a 4080 SUPER GPU, its high computational cost hinders the large-scale generation of high-quality challenging samples. To address these issues and to highlight the impact of hard samples on model performance, we propose a coarse-grained hard positive sample generation strategy.

By using a multimodal diffusion model\cite{croitoru2023diffusion}, we first segment a pedestrian image to obtain the clothing mask. Guided by a text prompt, the diffusion model then synthesizes new attire within the masked region:
\begin{equation}
    \mathbf{I}_{\text{new}} = \mathcal{G}(\mathbf{I}_{\text{src}} \cap M_{\text{cloth}}, t_{\text{prompt}})
\end{equation}
where $M_{\text{cloth}}$ denotes the clothing mask, $\mathcal{G}$ is the diffusion model, and $t_{prompt}$ is set to "generate a new cloth for the person." This process alters only the clothing area while preserving ID-specific features (e.g., facial appearance, body shape, or posture) and runs about 15 times faster than the try-on model.

It should be noted that this approach cannot reliably produce hard negative samples, as identical prompts still yield significant clothing variations. Nevertheless, it efficiently creates large volumes of hard positive samples without requiring labeled outfits. For models not dependent on clothing labels, these data serve effectively for pretraining, as shown in step 1 of Figure~\ref{frame}.

\subsection{Hard Sample Adaptive Learning}
\label{sec:adapt}

\subsubsection{\textbf{Hard Sample Analyzer (HSA)}}

Based on the definition of hard samples in Sec.~\ref{sec:hard_def}, we propose a Hard Sample Analyzer (HSA) to systematically identify hard samples across different datasets $\mathcal{D}$. Datasets encode clothing information in diverse ways: for instance, PRCC\cite{PRCC} distinguishes clothing based on indoor/outdoor settings, while LTCC\cite{LTCC} assigns a unique identifier to each pedestrian's clothing. During dataset loading, these representations are unified into independent clothing labels $c_i$ (e.g., 001, denoting "a person wearing clothing labeled [001]"), and camera labels are similarly mapped to $v_i$.

For an input batch of size $n$, we compute two $n \times n$ matrices: the Hard Positive Assess Matrix ($\text{IS\_HP}$) and the Hard Negative Assess Matrix ($\text{IS\_HN}$) to determine whether each sample pair qualifies as a hard sample. For batch $\mathcal{B} = \{\mathbf{x}_i\}_{i=1}^n$, constructing assessment matrices:
\begin{equation}
\small
\begin{aligned}
    &\mathrm{IS\_HP}_{ij} &= \mathbb{I}&[(y_i=y_j) \land (c_i\neq c_j \lor v_i\neq v_j)] \\
    &\mathrm{IS\_HN}_{ij} &= \mathbb{I}&[(y_i\neq y_j) \land (c_i = c_j)]
\end{aligned}
\end{equation}

\subsubsection{\textbf{Hard Sample Distance Adjustment (HSDA)}}

Metric learning is a common approach for feature learning in person ReID\cite{hermans2017defense}, aiming to minimize distances between anchors and positive samples while maximizing those with negative samples. However, models typically identify hard samples based on apparent similarity metrics, which may not align with the subjective definition of hard samples. By leveraging the Hard Sample Analyzer, truly hard samples can be identified for targeted supervised learning.

As shown in Figure~\ref{frame}, we propose the HSDA method to enhance the model's ability to distinguish hard samples by increasing the distance between hard positive pairs and decreasing it for hard negative pairs during training.

Specifically, we define two $N \times N$ matrices: the Hard Positive Matrix ($\text{HP\_M}$) and the Hard Negative Matrix ($\text{HN\_M}$). If $\text{IS\_HP}(i,j)$ is true, $\text{HP\_M}(i,j)$ is set to $1 + \alpha$; otherwise, it is $1$, increasing the distance between hard positive pairs. Similarly, if $\text{IS\_HN}(i,j)$ is true, $\text{HN\_M}(i,j)$ is set to $1 - \alpha$; otherwise, it is $1$, reducing the distance between hard negative pairs.
\begin{equation}
\text{HP\_M}(i, j) =
\begin{cases} 
1 + \alpha, & \text{if } \text{IS\_HP}(i, j) = 1, \\
1, & \text{otherwise},
\end{cases}
\end{equation}
\begin{equation}
\text{HN\_M}(i, j) =
\begin{cases} 
1 - \alpha, & \text{if } \text{IS\_HN}(i, j) = 1, \\
1, & \text{otherwise},
\end{cases}
\end{equation}

Thus, the distance between $i$ and $j$ in the feature space can be adjusted, enhancing the model's sensitivity to hard samples and increasing their learning intensity during loss optimization. This enables the model to focus on more critical features. The complete pipeline of our approach is presented in Algorithm~\ref{alg:overview}.

\textbf{An alternative perspective on HSDA} The proposed method can also be interpreted from the perspective of the loss function. Adjusting distances in the feature space corresponds to increasing the learning intensity during the loss computation. Simply put, $f(ax)$ and $af(x)$ yield the same gradient upon differentiation. Specifically, the standard triplet loss $\text{loss\_tri}$ is reformulated as:
\begin{equation}
\begin{split}
\mathcal{L}_{\text{HSDA}} = \max\Big( 0, & \ \text{HP\_M}(a, p) \cdot \text{D}(f_i^a, f_i^p) \\
& - \text{HN\_M}(a, n) \cdot \text{D}(f_i^a, f_i^n) + \text{margin} \Big)
\end{split}
\end{equation}
where $\text{D}(f_i^a, f_i^p)$ denotes the Euclidean distance between the anchor $a$ and the positive sample $p$, while $\text{D}(f_i^a, f_i^n)$ represents the Euclidean distance between the anchor $a$ and the negative sample $n$. By incorporating $\text{HP\_M}$ and $\text{HN\_M}$, the model could achieve more gradient information during training for optimization, enabling it to dynamically adjust the learning intensity for challenging positive and negative sample pairs with hard samples.

\begin{algorithm}[t]
\caption{HSGL Framework Workflow}
\label{alg:overview}
\begin{algorithmic}[1]
\State \textbf{Input:} Original dataset $\mathcal{D}$
\State \textbf{Output:} Trained ReID model $M_{\theta}$

\Procedure{Hard Sample Generation}{}
    \State \textit{\% Phase 1: Coarse-grained Generation (CHPG)}
    \State $\mathcal{M}_{\text{cloth}} \gets \Call{SegmentClothing}{\mathcal{D}}$
    \State $\mathcal{G}_{\text{coarse}} \gets$ 
    \Call{DiffusionGeneration}{$\mathcal{D}, \mathcal{M}_{\text{cloth}},$ "generate a new cloth for the person"}

    \State \textit{\% Phase 2: Fine-grained Generation (SHSG)}
    \State $\mathcal{D}_{\text{filter}} \gets \Call{FrontalPoseDetector}{\mathcal{D}}$
    \State $\mathcal{D}_{\text{highq}} \gets \Call{ImageQualityAssessor}{\mathcal{D}_{\text{filter}}}$

    \State $\mathcal{D}_{\text{topm}} \gets \Call{SelectTopM}{\mathcal{D}_{\text{highq}}, m}$
    
    \State $\mathcal{C}_{\text{clothes}}, \mathcal{T}_{\text{desc}} \gets$ 
    \Call{TryOffAnyone}{$\mathcal{D}_{\text{topm}}$}
    \State $\mathcal{D}_{\text{topn}} \gets \Call{SelectTopNPerID}{\mathcal{D}_{\text{highq}}, n}$
    \State $\mathcal{G}_{\text{fine}} \gets$
    \Call{MultimodalTryon}{$\mathcal{D}_{\text{topn}}, \mathcal{C}_{\text{clothes}}, \mathcal{T}_{\text{desc}}$}

    \State $\mathcal{D}_{\text{gen}} \gets \mathcal{G}_{\text{coarse}} \cup \mathcal{G}_{\text{fine}}$
\EndProcedure

\Procedure{Hard Sample Learning}{$\mathcal{D}, \mathcal{D}_{\text{gen}}$}
    \If{The base model is agnostic to clothing labels.}
        \State $M_{\theta} \gets \Call{Pretrain}{\mathcal{G}_{\text{coarse}}}$
    \EndIf

    \While{not converged}
        \State $\mathcal{B} \gets \Call{Sampling}{\mathcal{D} \cup \mathcal{G}_{\text{fine}}}$

        \State \textit{\% Hard Sample Analysis}
        \State $\text{IS\_HP}, \text{IS\_HN} \gets \Call{HardSampleAnalyzer}{\mathcal{B}}$
        \State $\text{HPA\_M} \gets \Call{ComputeHPMatrix}{\mathcal{B}, \text{IS\_HP}}$
        \State $\text{HNA\_M} \gets \Call{ComputeHNMatrix}{\mathcal{B}, \text{IS\_HN}}$

        \State \textit{\% Feature Learning}
        \State $\mathbf{F} \gets M_{\theta}(\mathcal{B})$
        \State $\mathbf{D} \gets \Call{PairwiseDistance}{\mathbf{F}}$

        \State \textit{\% Distance Adjustment}
        \State $\mathbf{D}_{\text{adj}} \gets \mathbf{D} \odot \text{HPA\_M} \odot \text{HNA\_M}$ 

        \State \textit{\% Loss Computation}
        \State $\mathcal{L}_{\text{cls}} \gets \Call{CrossEntropyLoss}{\mathbf{F}}$
        \State $\mathcal{L}_{\text{metric}} \gets \Call{AggregatedTripletLoss}{\mathbf{D}, \mathbf{D}_{\text{adj}}}$
        \State $\mathcal{L}_{\text{total}} \gets \mathcal{L}_{\text{cls}} + \lambda \mathcal{L}_{\text{metric}}$

        \State \Call{Backpropagate}{$\mathcal{L}_{\text{total}}$}
    \EndWhile
\EndProcedure
\end{algorithmic}
\end{algorithm}

\subsection{Loss Function}
\label{sec:loss}

\subsubsection{\textbf{ReID Loss}}

In cloth-changing scenarios, various technical approaches have been proposed\cite{hermans2017defense, CAL, AIM, FIRe}. The loss learning schemes from existing baseline models can be directly applied. A classic loss function in person re-identification is the cross-entropy classification loss, which maps feature vectors to identity labels. It is defined as:
\begin{equation}
\mathcal{L}_{cls}=-\frac{1}{B}\sum_{i=1}^B\sum_{c=1}^Cy_{i,c}\log(\hat{y}_{i,c})
\end{equation}
where $B$ is the batch size, $C$ is the number of identity classes, and $y_{i,c}$ is the ground-truth label for sample $i$ with respect to class $c$.

Our framework is compatible with any existing loss function methodologies. As a plug-and-play approach targeting hard sample learning, we recommend combining it with disentangled learning techniques for synergistic benefits. This combination enhances the model's ability to focus on identity-specific features while effectively handling challenging cases in cloth-changing scenarios.

\subsubsection{\textbf{Aggregated Triplet Loss}}

The proposed HSDA method operates as a metric learning technique in the feature space, with triplet loss serving as an effective and versatile mechanism\cite{hermans2017defense}. To preserve the model's inherent learning capabilities, we compute the triplet loss on both the original and adjusted distances, applying appropriate weighting to each. The complete loss function is formulated as follows:
\begin{equation}
\mathcal{L}_{\text{hatrip}} = \mathcal{L}_{\text{triplet}}(d) + 0.5 \cdot \mathcal{L}_{\text{triplet}}(d')
\end{equation}
in which the triplet loss is defined as:
\begin{equation}
\mathcal{L}_{\text{triplet}}(d) = \frac{1}{N}\sum_{i=1}^{N}\max(0, d(f_i^a, f_i^p) - d(f_i^a, f_i^n) + \text{margin})
\end{equation}

Here, $d$ represents the original distance metrics, and $d'$ denotes the distances adjusted using the hard sample matrices. This combination allows the model to focus on challenging samples while retaining its fundamental learning capabilities.

\subsubsection{\textbf{Overall Loss}}
The overall loss combines the baseline ReID loss with the adaptive triplet loss:
\begin{equation}
\small
\mathcal{L} = \mathcal{L}_{\text{ReID}} + \lambda\mathcal{L}_{\text{hatrip}}
\end{equation}
where $\lambda = 0.1$ for non-metric baselines and $\lambda_{\text{orig}}$ for metric-based methods. This design preserves baseline performance while enhancing the model's ability to learn from hard samples.

\begin{table}[h]
\centering
\setlength{\tabcolsep}{3.5pt} 
\caption{Performance comparison on two typical CC-ReID benchmark datasets PRCC and LTCC. † denotes reproduced results using provided open-source code.}
\begin{tabular}{c| c |c| c |c}
\hline
Dataset & Model & Venue & Rank-1 & mAP \\
\hline
\multirow{18}{*}{PRCC} 
    & RCSANet \cite{huang2021clothing} & ICCV '21 & 50.2 & 48.6 \\
    & TransReID \cite{TransReID} & ICCV '21 & 46.6 & 44.8 \\
    & MVSE \cite{MVSE} & ACM MM '22 & 47.4 & 52.5 \\
    & M2NET \cite{M2NET} & ACM MM '22 & 59.3 & 57.7 \\
    & CAL  \cite{CAL} & CVPR '22 & 55.2 & 55.8 \\
    & LDF \cite{chan2023learning} & TOMM '23 & 58.4 & 58.6 \\
    & AIM \cite{AIM} & CVPR '23 & 57.9 & 58.3 \\
    & CCFA \cite{CCFA} & CVPR '23 & 61.2 & 58.4 \\
    & SCNet \cite{SCNet} & ACM MM '23 & 59.9 & 61.3 \\
    & IGCL \cite{gao2023identity} & TPAMI '23 & 64.4 & 63.0 \\
    & CaAug \cite{liu2024cloth} & ACM MM '24 & 63.9 & 60.1 \\
    & CCPG \cite{pose_generation} & CVPR '24 & 61.8 & 58.3 \\
    & CLIP3DReID \cite{liu2024distilling} & CVPR '24 & 60.6 & 59.3 \\
    & MADE \cite{MADE} & TMM '24 & 64.3 & 59.1 \\

    & DLCR \cite{siddiqui2024dlcr} & WACV '24 & 66.5 & 63.0 \\
    & FIRe$^2$ (baseline) † \cite{FIRe} & TIFS '24 & 60.3 & 51.6 \\
    & IFD \cite{IFD} & ICASSP '25 & 65.3 & 61.7 \\
    \cline{2-5} 
    & \textbf{Ours} & - & \textbf{68.9} & \textbf{63.4} \\
\hline
\multirow{14}{*}{LTCC} 
    & TransReID \cite{TransReID} & ICCV '21 & 34.4 & 17.1 \\

    & 3DSL \cite{3DSL} & CVPR '21 & 31.2 & 14.8 \\
    & FASM \cite{FSAM} & CVPR '21 & 38.5 & 16.2 \\
    & CAL  \cite{CAL}  & CVPR '22 & 39.5 & 18.0 \\
    & Pos-Neg \cite{jia2022complementary} & TIP '22 & 36.2 & 14.4 \\
    & LDF \cite{chan2023learning} & TOMM '23 & 32.9 & 15.4 \\
    & 3DInvarReID \cite{3DInvar} & CVPR '23 & 37.8 & 16.7 \\
    & AIM \cite{AIM} & CVPR '23 & 40.6 & 19.1 \\
    & CCFA \cite{CCFA}& CVPR '23 & 45.3 & \textbf{22.1} \\
    & CLIP3DReID \cite{liu2024distilling} & CVPR '24 & 42.1 & 21.7 \\
    & DLCR \cite{siddiqui2024dlcr} & WACV '24 & 41.3 & 19.6 \\
    & FIRe$^2$ \cite{FIRe} & TIFS '24 & 44.6 & 19.1 \\
    & CSSC (baseline) † \cite{CSSC} & ICASSP '25 & 43.6 & 18.6 \\
    \cline{2-5} 
    & \textbf{Ours} & - & \textbf{45.4}& 19.2 \\
\hline
\end{tabular}
\label{performance}

\end{table}

\section{Experiments}
\

\textbf{Dataset and metrics.} We conducted experiments on two widely used clothing-changing person re-identification (CC-ReID) datasets: PRCC \cite{PRCC} and LTCC \cite{LTCC}. PRCC contains 33,698 images of 221 identities with two distinct clothing sets captured in indoor and outdoor conditions. LTCC includes 17,119 images of 152 identities, with an average of 3.1 clothing variations per identity. Performance was evaluated using Rank-1 accuracy and mean average precision (mAP).

\textbf{Implementation details.} We evaluated our hard sample strategy on four state-of-the-art models: CAL \cite{CAL}, AIM \cite{AIM}, FIRe$^2$ \cite{FIRe}, and CSSC \cite{CSSC}. For constructing the clothing library, we used TryOffAnyone \cite{xarchakos2024tryoffanyone} to extract high-quality clothing images matching the dataset styles. IDM-VTON \cite{choi2024improving} was employed for clothing changes in the pipeline, demonstrating high-quality performance.

For coarse-grained sample generation, we applied the SCHP \cite{li2020self} model for semantic segmentation to generate clothing masks, which were used as input for Stable Diffusion \cite{rombach2022high} to synthesize new clothing. These synthetic samples were used during pretraining. However, CAL \cite{CAL} and AIM \cite{AIM}, which rely on clothing labels, could not be pretrained with these unlabeled samples.

As the original FIRe$^2$  model underperformed on LTCC, we adopted its improved variant, CSSC \cite{CSSC}, for fairer comparisons. The coefficient $\lambda$ for the hard triplet loss ($\mathcal{L}_{\text{hatrip}}$) was set to $0.1$ for CAL and AIM, as their original implementations lack triplet loss. For FIRe$^2$ and CSSC, we retained the triplet loss coefficients from their original papers. Most experiments used a batch size of $32$, with other hyperparameters consistent with the baselines. All experiments were conducted on an NVIDIA RTX 4080 SUPER GPU.

\subsection{Comparison with the Baseline and State-of-the-art Methods}
We evaluated the baseline model's performance after integrating the proposed HSGL strategy and compared it with several state-of-the-art methods. As shown in Table~\ref{performance}, incorporating only $18\%\sim20\%$ additional high-quality hard samples and applying the HSAL strategy led to significant improvements of $3.1\%\sim4.9\%$ across various datasets and baselines. On the PRCC dataset \cite{PRCC}, combining coarse-grained sample pretraining with fine-tuning on high-quality hard samples set a new state-of-the-art (SOTA), improving Rank-1 accuracy and mAP by $8.9\%$ and $11.5\%$, respectively.

\subsection{Performance Improvement Curve}
\begin{figure}[t]
  \centering
  \includegraphics[width=\linewidth]{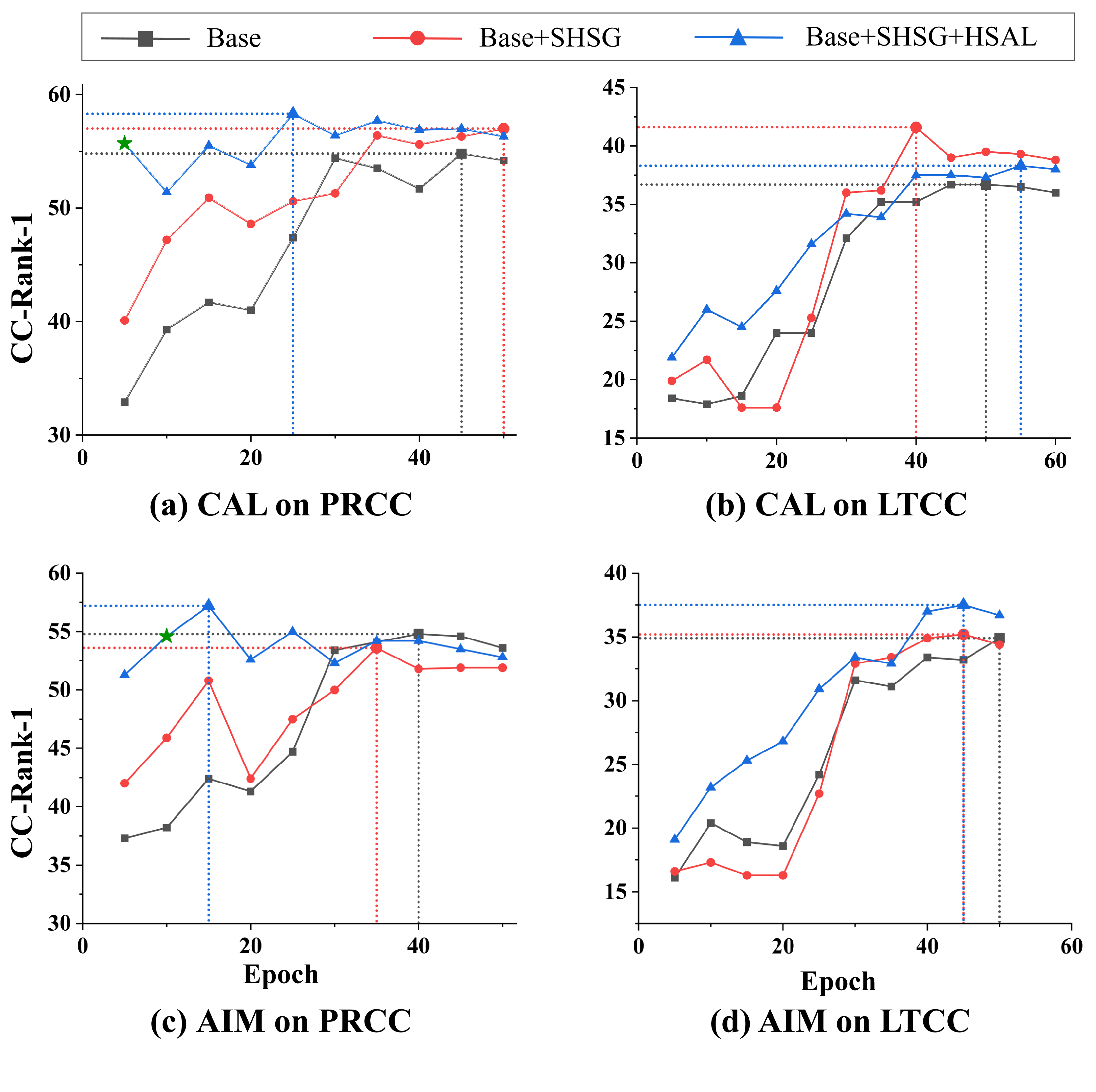}
  \caption{Accuracy improvement curves of CAL and AIM models on PRCC and LTCC datasets under different training strategies.
  }
  \label{accuracy_curve}
\end{figure}
By comparing the rate and magnitude of accuracy improvements across models, the benefits of hard sample generation and learning, particularly the contribution of HSAL are evident. With the addition of a small number of high-quality samples, overall model accuracy improves, as shown by the red curve in Figure~\ref{accuracy_curve}. This is expected, as more training samples enable the model to extract richer features within the same number of epochs.

The more significant improvement comes from HSAL, illustrated by the blue curve in Figure~\ref{accuracy_curve}. On the PRCC dataset, after applying the hard sample learning strategy, the model achieves baseline-level performance at the 50th/60th epoch by just the 5th/10th epoch, as marked by the green pentagram in the figure. This demonstrates that HSAL allows the model to capture more essential and discriminative features, resulting in faster convergence and higher accuracy.

\subsection{Ablation Study}

We conducted ablation studies to assess the individual and combined effects of hard sample generation and learning strategies, as shown in Table~\ref{ablation}.
\begin{table}[h]

\centering
\setlength{\tabcolsep}{4.5pt} 
\caption{Abalation study. The performance of ablated models evaluates the impact of combining each component on the PRCC cloth-changing scenario.}
\begin{tabular}{c| c c |c| c |c}
\hline
Baseline  & CHPSG & SHSG & HSAL & Rank-1 & mAP \\
\hline
\multirow{4}{*}{CAL} 
 &  \ding{55} &   \ding{55} & \ding{55} & 54.8 & 54.7 \\
 &  \ding{55} &   \ding{55} & \ding{51} & 54.8(\textcolor{red}{\textbf{+0.0}}) & 55.4 (\textcolor{red}{\textbf{+0.7}}) \\
 &  \ding{55} &   \ding{51} & \ding{55} & 57.7 (\textcolor{red}{\textbf{+2.9}}) & 55.7 (\textcolor{red}{\textbf{+1.0}})\\
 &  \ding{55} &   \ding{51} & \ding{51} & \textbf{61.2} (\textcolor{red}{\textbf{+6.4}}) & \textbf{60.7} (\textcolor{red}{\textbf{+6.0}}) \\
\hline
\multirow{4}{*}{AIM} 
 &  \ding{55} &   \ding{55} & \ding{55} & 54.8 & 55.7 \\
 &  \ding{55} &   \ding{55} & \ding{51} & 57.3(\textcolor{red}{\textbf{+2.5}}) & 57.9(\textcolor{red}{\textbf{+2.2}}) \\
 &  \ding{55} &   \ding{51} & \ding{55} & 53.6(\textcolor{green}{\textbf{-1.2}}) & 53.0(\textcolor{green}{\textbf{-2.7}}) \\
     &  \ding{55} &   \ding{51} & \ding{51} & \textbf{59.5} (\textcolor{red}{\textbf{+4.7}})  & \textbf{59.6}(\textcolor{red}{\textbf{+3.9}}) \\
 \hline
\multirow{7}{*}{FIRe$^2$} 
 &  \ding{55} &   \ding{55} & \ding{55} & 60.3 & 51.6 \\
 &  \ding{55} &   \ding{55} & \ding{51} & 60.0 (\textcolor{green}{\textbf{-0.3}}) & 51.9 (\textcolor{red}{\textbf{+0.3}}) \\
 &  \ding{55} &   \ding{51} & \ding{55} & 58.5(\textcolor{green}{\textbf{-1.8}}) & 47.7(\textcolor{green}{\textbf{-3.9}}) \\
 &  \ding{55} &   \ding{51} & \ding{51} & 60.9 (\textcolor{red}{\textbf{+0.6}}) & 52.9 (\textcolor{red}{\textbf{+1.3}})\\
 &  \ding{51} &   \ding{55} & \ding{55} & 64.0 (\textcolor{red}{\textbf{+3.7}}) & 56.5 (\textcolor{red}{\textbf{+4.9}}) \\
 &  \ding{51} &   \ding{51} & \ding{55} & 63.4 (\textcolor{red}{\textbf{+3.1}}) & 56.9 (\textcolor{red}{\textbf{+5.3}}) \\
 &  \ding{51} &   \ding{51} & \ding{51} & \textbf{68.9} (\textcolor{red}{\textbf{+8.6}}) & \textbf{63.4} (\textcolor{red}{\textbf{+11.8}}) \\
\hline
\end{tabular}

\label{ablation}
\end{table}

More interestingly, results indicate that using either strategy alone yields limited or even negative performance gains. For instance, applying SHSG on the CAL improved Rank-1/mAP by 2.5\%/1.0\%, but caused accuracy drops on AIM and FIRe$^2$, suggesting that without targeted learning, additional hard samples may confuse the model. Similarly, applying hard sample learning alone on the original dataset resulted in only minor improvements, likely due to the absence of sufficiently challenging samples. 

Furthermore, combining hard sample generation and learning significantly boosted performance across all datasets. Specific to CAL, Rank-1/mAP improved by 6.4\%/6.0\%, while on AIM increased by 4.7\%/3.9\%. On FIRe$^2$, our approach achieved new state-of-the-art results: 8.9\% in Rank-1 and 11.8\% in mAP.

These findings underscore the complementary nature of hard sample generation and learning. Generation enriches the dataset with critical but missing samples, while the learning strategy ensures their effective utilization. Together, they provide a robust solution for CC-ReID.

\subsection{Hyper-Parameter Sensitivity Analysis}

\begin{figure}[h]
  \centering
  \includegraphics[width=0.8\linewidth]{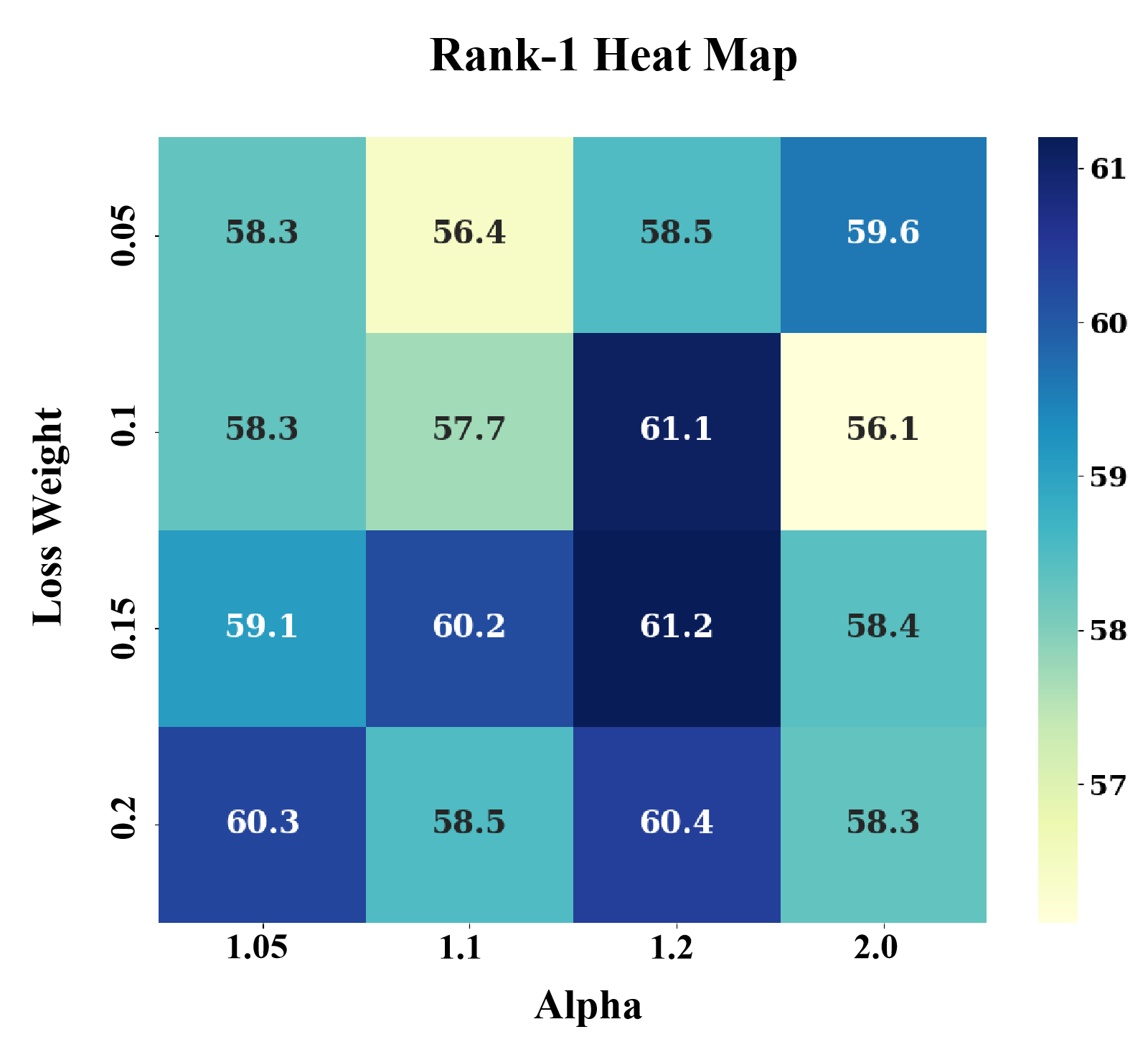}
  \caption{Rank-1 Accuracy Heatmap Based on the Distance Adjustment Factor ($\alpha$) and the Aggregated Triplet Loss Weight ($\lambda_{\text{hatrip}}$).
  }
  \label{heatmap}
\end{figure}

In the hard sample learning strategy, two key hyperparameters are involved: the weight of the aggregated triplet loss ($\lambda_{\text{hatrip}}$) and the distance adjustment factor ($\alpha$). A sensitivity analysis on the PRCC dataset using the CAL model, shown in Figure~\ref{heatmap}, reveals that appropriate tuning of these parameters generally improves performance. However, excessively high values lead to significant performance drops, likely due to overemphasis on hard samples, which causes the model to neglect general samples and compromise overall generalization.

\subsection{Visualization of Generated Samples}
\begin{figure}[t]
  \centering
  \includegraphics[width=\linewidth]{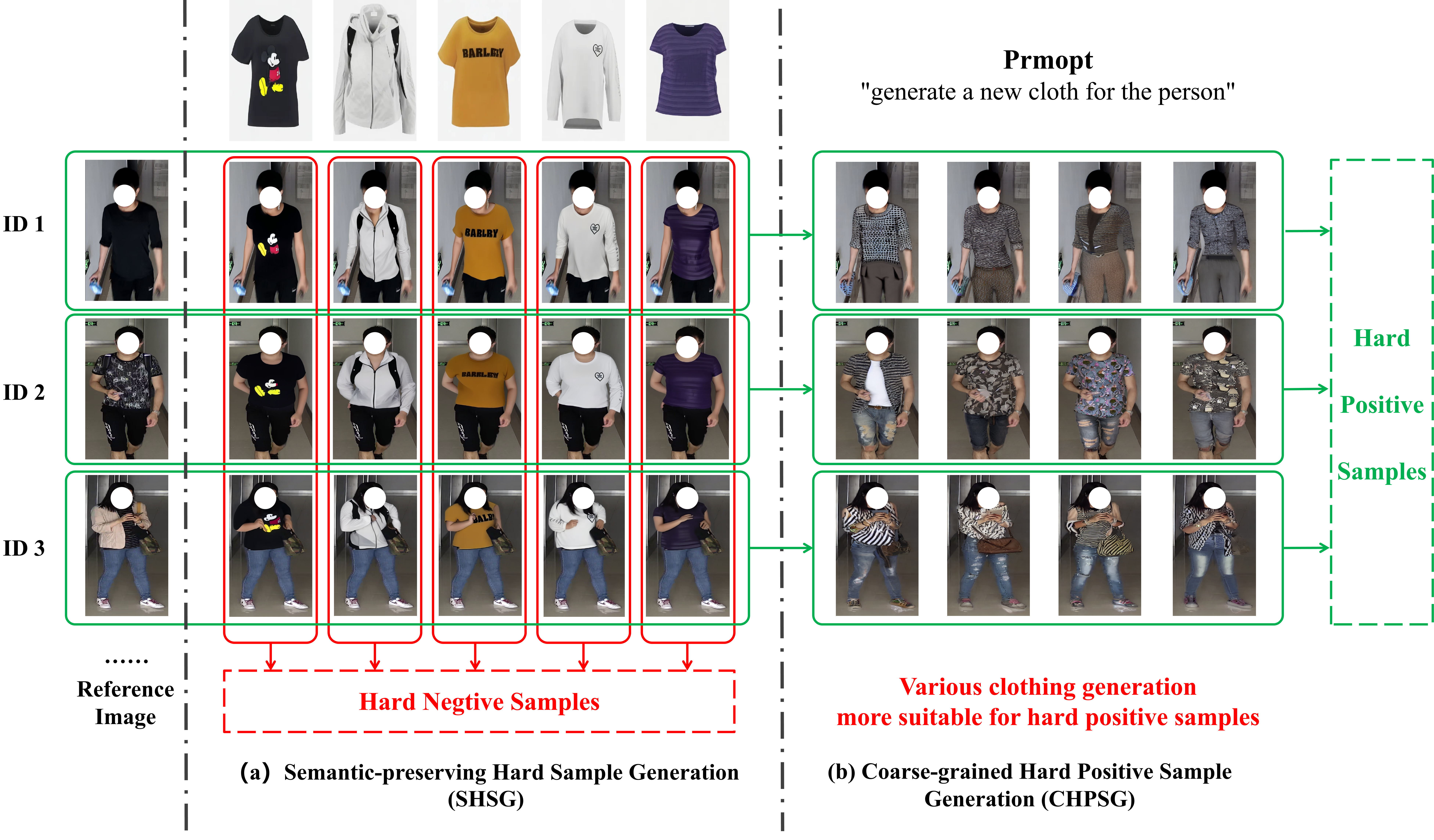}
  \caption{Samples generated by the two strategies of Dual-Granularity Hard Sample Generation (DGHSG).
  }
  \label{gen_vis}
\end{figure}
As shown in Figure~\ref{gen_vis}, SHSG generates pedestrian images with consistent clothing semantics, producing hard positive samples and hard negative samples in a controllable manner. CHPSG focuses on generating various outfits for the same identity, making it especially useful for creating hard positives. While it is less suited for reliable hard negatives due to occasional inconsistencies in clothing synthesis, it still contributes a large pool of diverse, open-domain negative samples. Overall, SHSG stands out in preserving identity and semantic alignment during clothing changes, enabling the generation of more precise and challenging hard samples. SHSG and CHPSG complement each other well, and their combination helps enrich the training data and support more targeted and effective model learning.

\section{Conclusion}

This paper proposes a novel multimodal-guided Hard Sample Generation and Learning (HSGL) framework to tackle the challenge of lacking effective targeted learning specific to significant hard samples in CC-ReID. By leveraging textual and visual modalities, we are the first to provide the explicit definition of hard samples for CC-ReID tasks. Furthermore, we achieve semantically consistent generation of hard positive and hard negative samples via our well-designed DGHSG strategy, which facilitates hard sample adaptive learning for improving the discrimination capability and robustness of the CC-ReID model. Experimental results illustrate that our HSAL framework achieves new SOTA performance on datasets PRCC and LTCC, meanwhile, significantly improving the training efficiency. These results highlight the effectiveness and generalization of our proposed model for CC-ReID tasks.


\clearpage
\bibliographystyle{ACM-Reference-Format}
\bibliography{sample-base}










\end{document}